\documentclass[
twocolumn,
hf,
]{ceurart}

\usepackage[numbers]{natbib}
\usepackage{appendix}
\usepackage{chngcntr}
\usepackage{float} 
\restylefloat{figure}
\usepackage[vlined,linesnumbered]{algorithm2e}
\SetKwInput{KwInput}{Input}
\usepackage{amsmath,amsfonts,amssymb,mathrsfs}
\usepackage{xcolor}
\usepackage{rotating}
\usepackage{booktabs} 
\usepackage{threeparttable}
\usepackage{subcaption}
\usepackage[normalem]{ulem}
\usepackage{mathrsfs}
\usepackage{newtxtext,newtxmath} 

\usepackage{amssymb}

\def\tsc#1{\csdef{#1}{\textsc{\lowercase{#1}}\xspace}}
\tsc{WGM}
\tsc{QE}
\tsc{EP}
\tsc{PMS}
\tsc{BEC}
\tsc{DE}

\begin{document}

\title [mode = title]{Testing Human-Hand Segmentation on In-Distribution and Out-of-Distribution Data in Human-Robot Interactions Using a Deep Ensemble Model}

\author[1,2]{Reza Jalayer}[type=editor,
   auid=000,bioid=1,
   orcid=0000-0003-3440-5658]
\ead{reza.jalayer@polimi.it}
\credit{We add the roles once we finish the paper}

\author[2]{Yuxin Chen}[type=editor,
   auid=000,bioid=1,
   orcid=0009-0002-0875-0343]
\credit{We add the roles once we finish the paper}

\author[3]{Masoud Jalayer}[type=editor,
   auid=000,bioid=1,
   orcid=0000-0001-8013-8613]
\credit{We add the roles once we finish the paper}

\author[1]{Carlotta Orsenigo}[type=editor,
   auid=000,bioid=1,
   orcid=0000-0001-8688-414X]
\credit{We add the roles once we finish the paper}

\author[2]{Masayoshi Tomizuka}[
   auid=000,bioid=1,
   orcid=0000-0003-0206-6639]
\credit{We add the roles once we finish the paper}

\address[1]{Department of Management, Economics and Industrial Engineering, Politecnico di Milano, Via Lambruschini 24/b, 20156, Milan, Italy}
\address[2]{Department of Mechanical Engineering, University of California at Berkeley, Berkeley, CA 94709, USA}
\address[3]{Department of Materials and Mechanical Engineering, University of Turku, Vesilinnantie 5, Turku, 20014, Finland}


\conference{ArXiv}

\begin{abstract}
Reliable detection and segmentation of human hands are critical for enhancing safety and facilitating advanced interactions in human-robot collaboration. Current research predominantly evaluates hand segmentation under in-distribution (ID) data, which reflects the training data of deep learning (DL) models. However, this approach fails to address out-of-distribution (OOD) scenarios that often arise in real-world human-robot interactions.
In this study, we present a novel approach by evaluating the performance of pre-trained DL models under both in-distribution (ID) data and more challenging out-of-distribution (OOD) scenarios. To mimic realistic industrial scenarios, we designed a diverse dataset featuring simple and cluttered backgrounds with industrial tools, varying numbers of hands (0 to 4), and hands with and without gloves. For OOD scenarios, we incorporated unique and rare conditions such as finger-crossing gestures and motion blur from fast-moving hands, addressing both epistemic and aleatoric uncertainties. 
To ensure multiple point of views (PoVs), we utilized both egocentric cameras, mounted on the operator's head, and static cameras to capture RGB images of human-robot interactions. This approach allowed us to account for multiple camera perspectives while also evaluating the performance of models trained on existing egocentric datasets, such as EgoHands, and Ego2Hands, as well as static-camera datasets, including HAGS and HADR.
For segmentation, we used a deep ensemble model composed of UNet and RefineNet as base learners. Performance evaluation was conducted using segmentation metrics and uncertainty quantification via predictive entropy. Results revealed that models trained on industrial datasets (HADR and HAGS) outperformed those trained on non-industrial datasets (EgoHands and Ego2Hands), highlighting the importance of context-specific training. Although all models struggled with OOD scenarios, those trained on industrial datasets demonstrated significantly better generalization.

\end{abstract}


\maketitle


\section{Introduction}

In the era of Industry 4.0, industrial robots and humans increasingly work side-by-side in collaborative environments, moving away from the traditional approach of isolating robots within cages to prevent accidents \citep{grau2020robots,tantawi2019advances}. This shift underscores the critical importance of human safety, particularly in preventing or mitigating collisions in shared workspaces \cite{kim2022collision}. Hands are among the most vulnerable parts of the human body in these interactions which are constantly present in the proximity of industrial robots \cite{benmessabih2024online}. Reliable detection and segmentation of human hands are therefore essential to ensure human safety as a main objective. Beyond safety—one of the cornerstones of Industry 4.0—Industry 5.0 expands the focus to emphasize human-centric interactions. This newer approach encourages manufacturers to prioritize not only safety but also the comfort and well-being of operators during human-robot interactions \cite{alves2023industry}. In this context, enabling robots to understand human hand gestures and actions through accurate hand segmentation has become an area of growing research interest \cite{qi2024computer}.

There are various approaches to detecting human hands around industrial robots, ranging from wearable sensors to computer vision-based methods. Wearable sensors, such as gloves equipped with tracking devices \cite{gao2021dynamic,huang2022virtual} or markers for positional detection \cite{tamantini2021wgd}, have been explored in previous studies. However, the constant need to wear these devices during operation can be frustrating for workers, and the setup requirements before use further limit their practicality in industrial environments. In contrast, computer vision-based techniques offer a more operator-friendly solution, allowing seamless hand detection and segmentation without requiring workers to wear any additional equipment. The rapid advances in Machine Learning (ML) and, more specifically, Deep Learning (DL), coupled with the growing availability of powerful hardware, have further propelled computer vision as the preferred choice for hand detection and segmentation. Unlike traditional computer vision methods, such as template matching \cite{yun2012hand}, which demand manual intervention and yield less accurate results, DL-based models deliver superior accuracy and robustness, making them more suitable for complex industrial applications \cite{qi2024computer}.

DL-based hand segmentation models heavily depend on the datasets they are trained on \cite{taran2024benchmarking}. While several hand segmentation datasets have been introduced in the literature and made publicly accessible, significant challenges arise when applying these models to human-robot interactions in industrial domains. One primary challenge is that the backgrounds in these datasets are often biased toward accessible environments, such as residential settings or scenes containing common household objects \cite{sharma2024collection}. This bias makes it difficult for models trained on such datasets to generalize to industrial scenarios with complex and cluttered backgrounds. Another challenge lies in the limited number of hands present in most datasets, typically restricted to one or two per scene. However, in real-world interactions with robots, multiple participants or hands may simultaneously be in the frame. Furthermore, these datasets often capture hand images using either egocentric and mobile or static cameras, while real-world scenarios may involve different camera setups that deviate from these idealized conditions.
Additionally, the gestures captured in these datasets are often casual and fall within predefined categories, overlooking the diversity and complexity of hand gestures in actual industrial interactions. For instance, gestures such as finger crossings or interactions with robotic hands are rarely represented. Moreover, most datasets predominantly feature bare human hands, ignoring the fact that industrial operators typically wear gloves for safety, which alters the appearance and texture of hands \cite{sharma2024collection}.

On the other hand, creating new datasets that address these challenges is a labor-intensive process, requiring significant time and resources. Even if such datasets were created, their generalization to other human-robot tasks and conditions would remain uncertain. Consequently, there is a notable gap in the literature exploring how well pre-trained models, trained on existing datasets, perform when applied to real-world industrial hand segmentation tasks.

In this study, we address these challenges by conducting a novel investigation into the performance of DL models for hand segmentation under both in-distribution (ID) and out-of-distribution (OOD) conditions. We train models using non-industrial datasets, such as EgoHands\cite{bambach2015lending} and \cite{lin2020ego2hands}, as well as an industrial-like dataset, HAGS and HADR \cite{sharma2024collection,grushko2023hadr}. To evaluate these models, we collect real-world images of human-robot interactions in industrial environments, encompassing a wide range of scenarios. These include ID conditions that align with the training datasets and OOD conditions that introduce novel challenges, such as complex gestures.

To ensure accurate evaluations, we use a deep ensemble model, an uncertainty-aware DL approach \cite{abdar2021review}. By ensembling segmentation models that have been successfully used in the field, we create a deep ensemble model to quantify model uncertainty on both ID and OOD data as well as assessing the accuracy of hand segmentation.

The remainder of this paper is structured as follows: In Section~\ref{sec:related_works}, we provide a detailed overview of related works, highlighting previous approaches, datasets, and the challenges of applying existing methods to industrial human-robot interactions. Section~\ref{sec:datasets} describes the datasets and experimental setup, detailing the training datasets, test data collection, and experimental methodology, including details of segmentation models and metrics. In Section~\ref{sec:results}, we present the results of our experiments, analyzing model performance on different data conditions, insights from segmentation performance and uncertainty quantification as well as qualitative results. Finally, in Section~\ref{sec:conclusions}, we summarize our findings and propose directions for future research.

\section{Related works}
\label{sec:related_works}
Hand segmentation have gained significant attention in recent years, particularly in computer vision-based applications. These techniques have been applied to a variety of domains, including medical purposes such as rehabilitation tasks \cite{dutta2023efficient} and sign language detection for speech-impaired individuals \cite{bandini2020analysis, rastgoo2020hand}. They are also widely used in human-computer interaction scenarios, especially with the advancement of augmented reality (AR), virtual reality (VR), and mixed reality (MR) technologies \cite{wu2023real, hassan20233d}. Within the context of human-robot interaction (HRI), hand segmentation based on computer vision techniques has facilitated applications such as gesture recognition for controlling mobile robots \cite{sahoo2023hand,xu2020skeleton,gao2019dual} and assisting surgical robots \cite{van2021gesture}. Despite this progress, the application of hand segmentation in industrial HRI has received comparatively less attention, where unique challenges demand tailored approaches.

While many recent works employ DL models for hand segmentation, these models face limitations when applied to industrial settings. There are few and recent works in human-robot interactions using hand segmentations. For instance, Sajedi et al. \cite{sajedi2022uncertainty} leveraged a Bayesian Neural Network for hand segmentation, focusing on RGB images captured from a third-person perspective who carries a mobile phone to record the data in a human-robot interaction. Their model, trained on the EgoHands dataset \cite{bambach2015lending}, incorporated uncertainty quantification to improve segmentation accuracy. The hand instances in their data were restricted to two hands of the operator working with the robot. Also, the experimental context could be closer to reality by the presence of gloves or cluttered workspaces. Grushko et al. \cite{grushko2023hadr} introduced HADR, a synthetic RGB-D dataset designed for industrial applications, using domain randomization to mimic real-world conditions. Despite its contributions, HADR is limited by its static top-down camera angles and a restriction to two hand instances per frame, which fails to capture the dynamic and multi-user nature of industrial collaboration. Sharma et al. \cite{sharma2024collection} recently addressed some of these gaps by introducing the HAGS dataset, which includes gloved and ungloved hands captured from stationary side and top-down cameras. However, the dataset also limits hand instances to two per frame and does not explore rare gestures or the influence of background complexity. Their results with segmentations with DL-based models e.g. UNet trained on the available non-industrial datasets resulted in non-satisfactory results emphasizing the importance of creating more domain-specific datasets like HAGS for future research.

Based on these works, several critical limitations and challenges emerge. First, existing datasets for hand segmentation in HRI restrict the number of hands to two, overlooking multi-user industrial scenarios where more than two hands may be present. Second, the gestures captured in these datasets are often casual and lack diversity, failing to account for rare but practical gestures such as interlocked fingers or crossed hands. Third, no studies considered noisy conditions such as motion blur, which often occur during dynamic interactions with robots. Additionally, while gloves are commonly used in industrial settings, only HAGS, includes gloved hands. Finally, existing studies predominantly rely on static cameras, with no research on considering both egocentric and static cameras to provide a better evaluation.

Despite these limitations, creating new datasets tailored to industrial HRI remains a labor-intensive task, and questions about their generalization to diverse scenarios persist. There is also a lack of studies evaluating the performance of pre-trained models in real-world industrial conditions. Moreover, the relationship between segmentation accuracy and background complexity (e.g., cluttered environments vs. object-free scenes) has been largely unexplored in the literature, which limits our understanding of the robustness of existing models.

Our study addresses these gaps by testing the performance of state-of-the-art DL models on a test dataset that incorporates both in-distribution (ID) and out-of-distribution (OOD) conditions. This work examines challenging scenarios, including rare gestures, motion blur, gloved and non-gloved hands across a combination of egocentric and static camera perspectives. By leveraging uncertainty-aware deep ensemble models, we aim to provide an evaluation of hand segmentation performance and contribute insights for future advancements in industrial human-robot interactions.

\section{Datasets and experimental setup}
\label{sec:datasets}

This section provides details on the datasets used for training and testing the deep learning models, as well as the experimental setup designed to evaluate their performance. We describe the datasets we choose to train our models, our custom-designed test dataset for in-distribution (ID) and out-of-distribution (OOD) evaluation, and the methodology and pipeline of our work.

\subsection{Training datasets}

For training the deep learning (DL) models, we considered both industrial and non-industrial datasets. This was done because on one hand there is no hand segmentation dataset with egocentric view in industrial-like context and on the other considering trained model on industrial and non-industrial images gives us a better evaluation for our results. The summary of the description of each training dataset is provided in Table\ref{tab:1}. 

\begin{table*}[]
\caption{Summary of four training datasets.}
\resizebox{\textwidth}{!}{
\centering
\label{tab:1}
\begin{tabular}{lcccccc}
\toprule
Training dataset & Point of View            & Industrial & Real/Synthetic & RGB/RGB-D & Max No. of hands per frame & No. of annotaded frames \\ \midrule
EgoHands \cite{bambach2015lending}        & Egocentric      & No         & Real           & RGB       & 4                          & 4800                    \\ 
Ego2Hands \cite{lin2020ego2hands}       & Egocentric      & No         & Real           & RGB       & 2                          & 2000                    \\ 
HADR \cite{grushko2023hadr}            & Top-down        & Yes        & Synthetic      & RGB-D     & 2                          & 117000                  \\ 
HAGS \cite{sharma2024collection}            & Top-Down + Side & Yes        & Real           & RGB       & 2                          & 1728                    \\ 
\bottomrule
\end{tabular}
}
\end{table*}

Since no industrial-like datasets exist for hand segmentation in egocentric views, we utilized two widely recognized non-industrial datasets: EgoHands and Ego2Hands. EgoHands is a very large available egocentric dataset for hand segmentation, comprises 4,800 pixel-level annotated frames captured using Google Glass. The frames depict diverse non-industrial backgrounds and contain between zero to four hands per image. This dataset has been widely employed for training hand segmentation models in a variety of applications, including human-robot interaction. However, it lacks relevance to industrial contexts, such as the use of gloves or cluttered backgrounds, which limits its applicability to more complex environments.
Ego2Hands is a more recent egocentric dataset with 2,000 annotated frames of RGB images, each containing a single person and up to two hand instances. This dataset employs background replacement techniques, providing greater control over environmental diversity compared to EgoHands. Additionally, it introduces inter-occluded hands for the first time and ensures a more even spatial distribution of hand positions across frames, addressing some limitations of its predecessor. Despite these advancements, Ego2Hands remains focused on non-industrial settings, with backgrounds and hand interactions that do not reflect the complexities of industrial human-robot interaction scenarios.

To incorporate industrial perspectives, we included two datasets designed specifically for such contexts: HADR and HAGS. HADR is a synthetic dataset with RGB-D annotations, created using domain randomization techniques to reduce the reality gap between synthetic and real-world data. This was achieved by introducing random variations in distractor object properties, background textures, camera positions and orientations, and lighting conditions. The dataset contains 117,000 annotated frames, making it one of the largest datasets available for hand segmentation. For our study, we used only the RGB modality of HADR, as our test data, described later, was captured using RGB cameras. Each frame in HADR contains a maximum of two hand instances, all viewed from a top-down perspective. While the dataset is comprehensive in terms of synthetic diversity, its static camera viewpoint and synthetic nature limit its application in fully dynamic or real-world settings.
HAGS, on the other hand, provides a realistic dataset consisting of 1,728 frames with pixel-level annotations of human hands interacting with industrial robots. The dataset includes both gloved and ungloved hands, addressing a key limitation of many previous datasets. Images in HAGS were captured using two stationary cameras: a GoPro Hero 7 capturing top-down views and a RealSense Development Kit Camera SR300 capturing side views. To increase variability, the dataset incorporated background replacement techniques, using both real and synthetic backgrounds. However, like HADR, HAGS restricts the number of hands per frame to two instances and focuses on industrial scenarios while overlooking the presence of some challenging conditions like rare hand gestures.

\subsection{Test dataset design}

To evaluate the trained models, we created a realistic image dataset by recording interactions with an industrial robotic arm (FANUC LR Mate 200iD 7L robot). The dataset was captured from two camera perspectives: a static side-view camera and an egocentric camera mounted on the operator's head. This dual-camera setup was chosen to provide comprehensive coverage of the interaction scene and to evaluate model performance across varying viewpoints. 

\begin{figure*}[ht]
 \centering
 \includegraphics[width=0.8\linewidth]{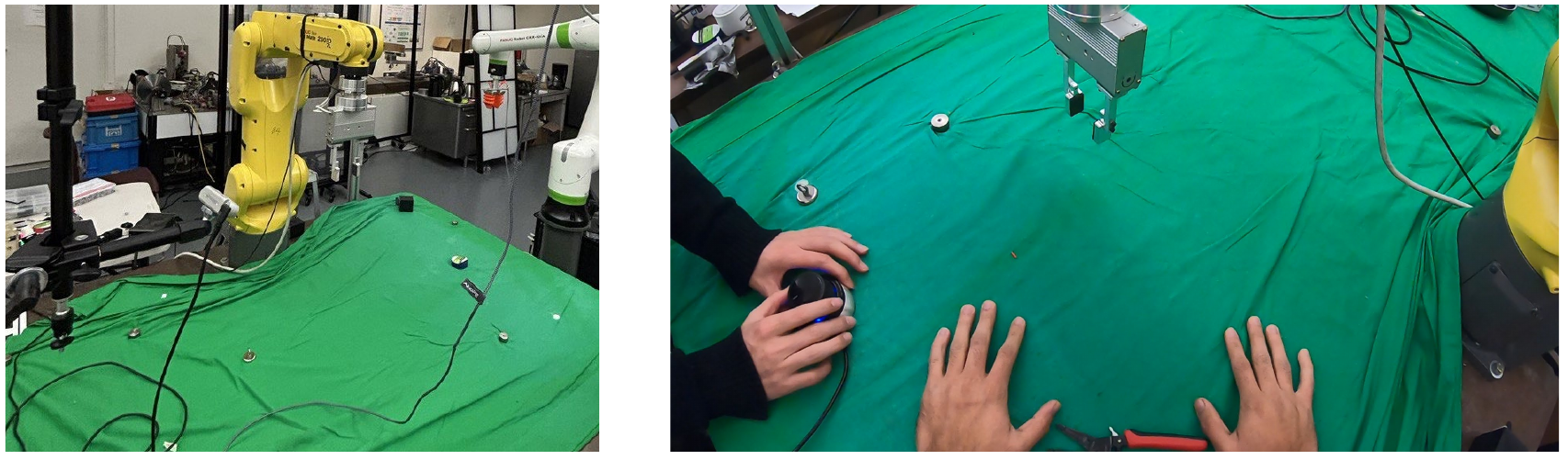}
 \caption{Dataset collection setup. We mount an Intel RealSense D435 on the right-hand side of the robotic arm for static images (left), and a GoPro camera on the helmet of the operator to capture egocentric images (right).}
 \label{fig:1}
\end{figure*}

For the side view, we used an Intel RealSense D435 camera, positioned on the right-hand side of the robot, as shown in the left of Figure \ref{fig:1}. The camera angle was carefully adjusted to capture the interaction space, including the operator’s hands and the robotic arm, ensuring that the entire workspace was visible. For the egocentric view, a GoPro camera was mounted on the operator’s helmet using a headband, allowing us to capture the perspective of the operator during the interaction as evident in the left picture of Figure \ref{fig:1}.

The dataset includes diverse interaction scenarios with one and two operators closely working with the robot. To ensure varied background complexity, we recorded videos in both object-free environments and cluttered industrial settings with industrial tools (e.g. hammer, scissors, wrenches, nuts and bolts, ...) present. Hands were captured with and without gloves to reflect realistic industrial conditions, where operators frequently wear gloves for safety. Additionally, some videos intentionally included rare hand gestures (e.g., interlocked fingers and crossed hands) to create out-of-distribution (OOD) data for testing in these conditions. Motion-blurred frames caused by fast-moving hands were also included to simulate real-world challenges as aleatoric uncertainty.

\begin{figure*}[ht]
 \centering
 \includegraphics[width=0.9\linewidth]{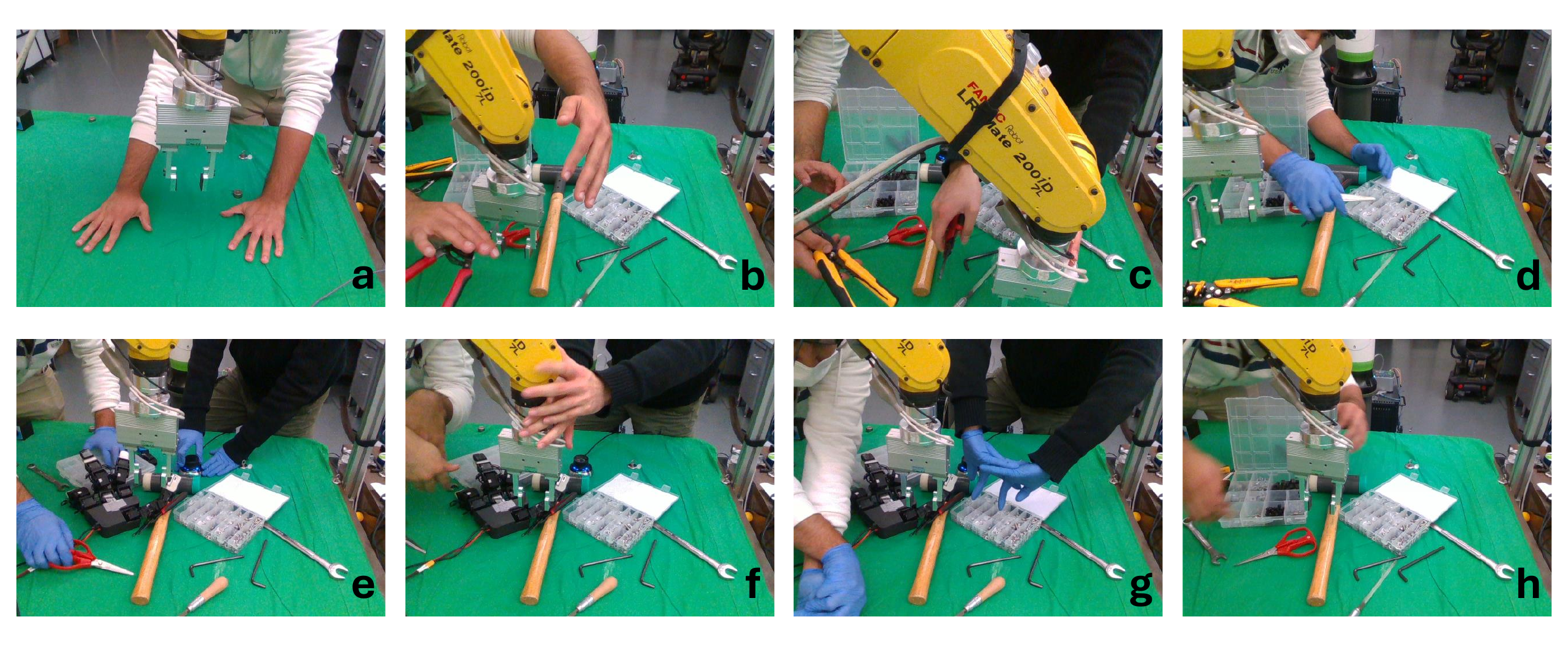}
 \caption{Different scenarios of human-robot interactions from the side view static camera:  one operator (O1) in (a) simple background and (b) cluttered background, (c) presence of two operators (O2), (d) one operator (O1) with gloves (GH), (e) two operators (O2) with gloves (GH), (f) two operators (O2) with rare gestures (RG), (g) two operators (O2) with gloves (GH) having rare gestures (RG), and (h) image with motion blurred noise (MBN).}
 \label{fig:2}
\end{figure*}

To streamline the description of dataset conditions, we use the following abbreviations: one operator (O1), two operators (O2), gloved hands (GH), hands with rare gestures (RG), and motion-blurred noisy hands (MBN). Figure \ref{fig:2} illustrates these conditions for clarity.

\subsection{In-distribution (ID) and out-of-distribution (OOD) scenarios}

For effective evaluation of deep learning (DL) models, it is crucial to assess their performance not only on data that aligns with their training distribution (in-distribution (ID) data) but also on data that deviates from this distribution (out-of-distribution (OOD) data). The discrepancy between ID and OOD data introduces uncertainty in the model's predictions, which can significantly affect their reliability. Understanding and addressing this uncertainty is essential, especially in safety-critical applications like human-robot interaction.
Uncertainty in predictions can arise from two main sources: epistemic uncertainty and aleatoric uncertainty \cite{ovadia2019can}. Epistemic uncertainty, often referred to as knowledge uncertainty, occurs when the model has not encountered all the characteristics of the data during training. This type of uncertainty can be reduced by introducing additional training data that better represents these unseen scenarios. On the other hand, aleatoric uncertainty, also known as data uncertainty, arises due to inherent variability in the data, such as noise or measurement errors. Unlike epistemic uncertainty, aleatoric uncertainty is irreducible and cannot be mitigated by adding more training data \cite{abdar2021review}.

In our experiments, OOD data stemming from epistemic uncertainty includes conditions absent in the training datasets, such as gloved hands (GH), rare gestures (RG), and the presence of two human operators (O2), when the training data contained only one operator. Aleatoric uncertainty arises from noisy conditions, such as motion-blurred hands (MBN), which were not included in the training datasets. These conditions were specifically designed to evaluate the robustness and generalization capabilities of the models, as summarized in Table \ref{tab:2}.

\begin{table*}[]
\caption{OOD scenarios of the training dataset.}
\resizebox{\textwidth}{!}{
\centering
\label{tab:2}
\begin{tabular}{lcl}
\toprule
\multirow{2}{*}{Training dataset} & \multicolumn{2}{c}{OOD Scenarios}                                                                                      \\ \cline{2-3} 
                                  & Epistemic Uncertainty                                           & \multicolumn{1}{c}{Aleatoric Uncertainty}            \\ 
                                  \midrule
EgoHands \cite{bambach2015lending}                          & gloved hands (GH), rare gestures (RG)                           & \multicolumn{1}{c}{motion-blurred noisy hands (MBN)} \\ 
Ego2Hands \cite{lin2020ego2hands}                        & two human operators (O2), gloved hands (GH), rare gestures (RG) & motion-blurred noisy hands (MBN)                     \\ 
HADR \cite{grushko2023hadr}                             & two human operators (O2), gloved hands (GH), rare gestures (RG) & motion-blurred noisy hands (MBN)                     \\ 
HAGS \cite{sharma2024collection}                             & two human operators (O2), rare gestures (RG)                    & motion-blurred noisy hands (MBN)                     \\ 
\bottomrule
\end{tabular}
}

\end{table*}


As shown in Table \ref{tab:2}, certain scenarios are consistently classified as OOD across all training datasets. For instance, rare gestures (RG) are absent in all datasets, making them OOD for all trained models. Similarly, motion-blurred hands (MBN), representing aleatoric uncertainty, are not included in any training datasets and are treated as OOD for all models.
However, some scenarios depend on the specific training dataset. For example, gloved hands (GH) are considered ID for models trained on HAGS, as this dataset includes gloved hands during training, but OOD for models trained on EgoHands, Ego2Hands, and HADR. Likewise, two human operators (O2) are considered ID for models trained on EgoHands, as this dataset includes scenes with multiple operators, but OOD for other datasets where the presence of two operators is absent.

By generating diverse ID and OOD scenarios, our framework systematically evaluates the robustness and generalization capabilities of the trained models.

\subsection{Data preparation}


To construct the test dataset, we recorded nine videos capturing interactions between one and two operators with an industrial robotic arm, using both side-view and egocentric cameras. In total, 34,577 frames were recorded from each camera and each frame for the side camera has the dimensions of 640 x 480 pixels and for the egocentric camera 1920 x 1080 pixels. Since the videos were captured using high-frame-rate cameras, many frames were repetitive, with minimal or no change between consecutive frames, making them redundant for our analysis. To address this, we conducted a manual review of the recorded videos, carefully selecting frames that were non-repetitive and meaningful for our study. After this refinement process, 1,871 unique frames were retained from each camera (side-view and egocentric).

For the testing phase, we aimed to achieve a balanced dataset across the 8 conditions defined in Figure \ref{fig:2}. From each condition, 20 frames were randomly selected for annotation, resulting in a total of 320 annotated images (160 frames each for the side and egocentric views).

\begin{figure*}[ht]
 \centering
 \includegraphics[width=0.6\linewidth]{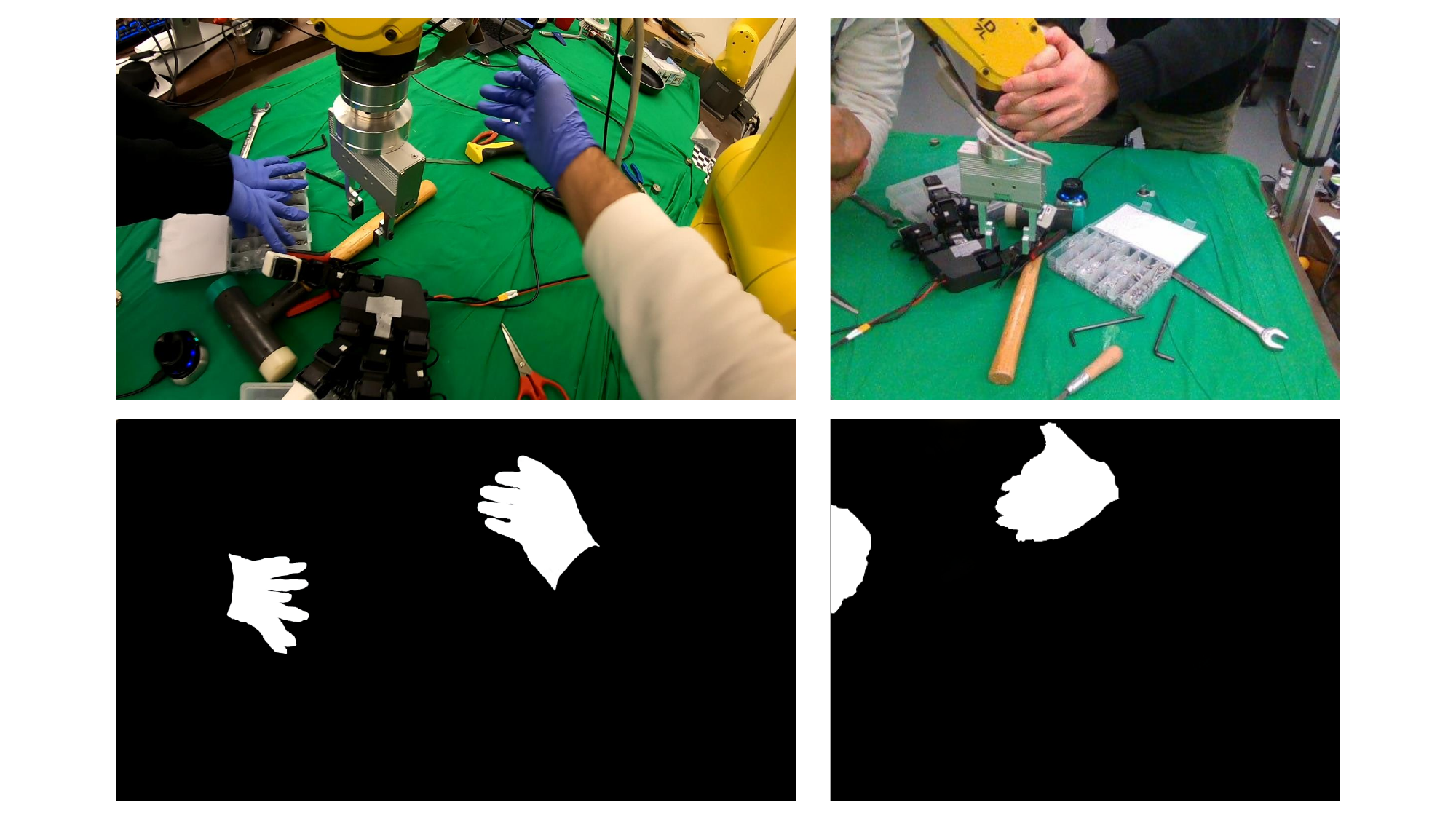}
 \caption{Sample frames with ground truth masks, captured from egocentric view (left) and side view camera (right).}
 \label{fig:3}
\end{figure*}

The selected images were annotated with pixel-wise accuracy for binary classification, where hand regions were labeled as the foreground and all other areas as the background. To achieve high-quality annotations, we utilized Label Studio \cite{tkachenko2020label}, an open-source data labeling tool, to create initial coarse annotations. These annotations were meticulously refined manually to ensure precision and consistency across the dataset. As the trained models are designed to segment only the hand region—from the tips of the fingers to the wrist—we adhered to this definition, excluding the arm and forearm from the annotated hand regions to maintain alignment with the training datasets. Figure \ref{fig:3} illustrates examples of annotated images, showcasing the accuracy and uniformity of the labeling process.

This rigorous data preparation workflow resulted in a dataset that is not only diverse and representative of various conditions but also meticulously balanced across all categories. By ensuring high annotation quality and meaningful representation of scenarios, the dataset is well-suited for comprehensively evaluating the performance of DL models under different conditions.

\subsection{Models and metrics}

In this work, we used deep ensemble for the evaluations. Deep ensembles are consist of multiple of DL models called base learners which are trained independently to improve predictive accuracy and make the uncertianty of predictions quantifiable \cite{abdar2021review}. For the choice of base learners, we selected two widely recognized DL algorithms for segmentation: UNet \cite{ronneberger2015u} and RefineNet \cite{lin2017refinenet}. These models leveraged the encoder-decoder technique and have been successfully employed in prior works on human hand segmentation \cite{dutta2020semantic, sharma2024collection, lin2020ego2hands, urooj2018analysis}.

\begin{figure*}[ht]
 \centering
 \includegraphics[width=12cm]{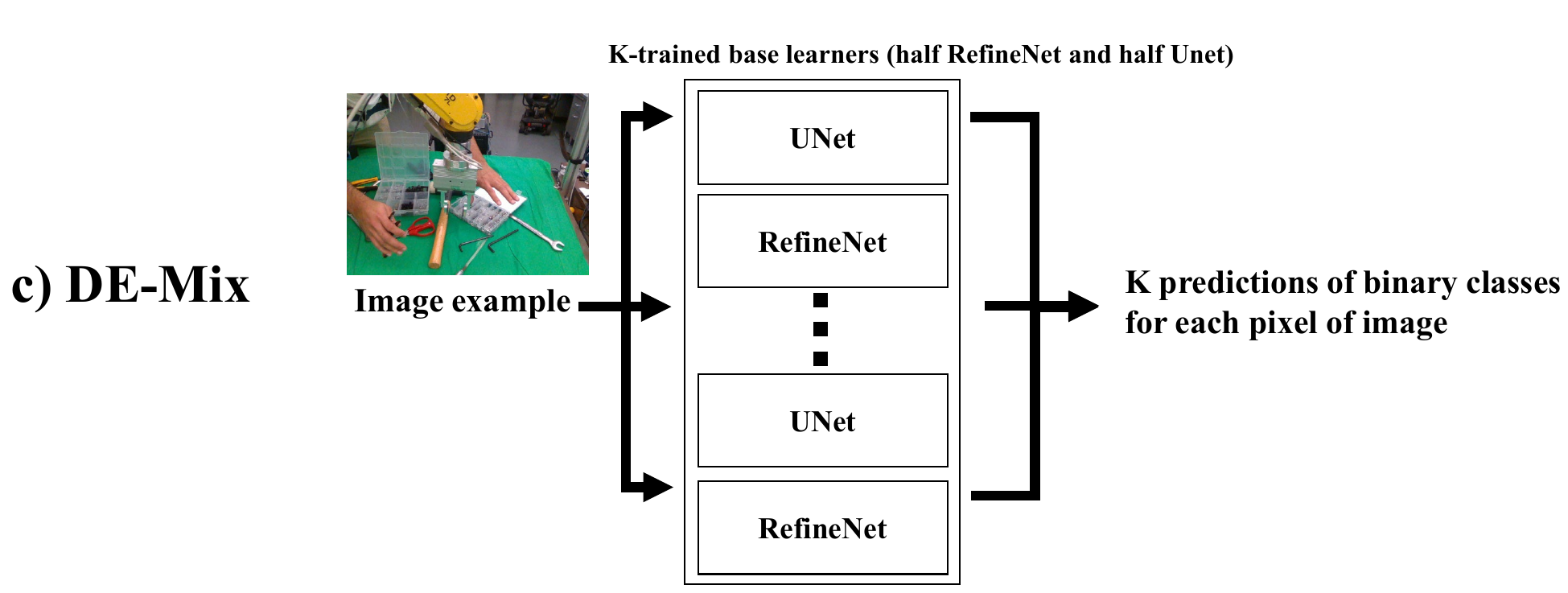}
 \caption{Deep ensemble architectures (DE-Mix) made by K-trained models which half are UNet and RefineNet.}
 \label{fig:4}
\end{figure*}

The configuration of this model is illustrated in Figure \ref{fig:4}. in this configuration, DE-Mix, combines both UNet and RefineNet models as base learners. This approach incorporates the concepts of the heterogeneous ensembles (different base learners), as discussed in prior ensemble learning research \cite{ganaie2022ensemble}. Using heterogeneous ensembles, have been shown to enhance performance by leveraging the diversity of base learners \cite{han2022out}. While deep ensemble models have been applied to hand segmentation tasks before \cite{sharma2024collection}, to the best of our knowledge, the prior work has employed only homogenous ensembles combining identical DL architectures for hand segmentation, such as ensemble of UNet \citep{sharma2024collection}.

As shown in Figure \ref{fig:4}, the output of the deep ensemble is derived from the predictions of K-base learners. Since we do binary classification of segmenting hands from background, each base learner performs binary pixel-wise classification on the input image, determining whether each pixel belongs to the background or a hand. The final output of the deep ensemble is computed as the average prediction across the K-base learners.

To evaluate segmentation performance, we use the mean Intersection over Union (mIoU), a widely accepted metric that quantifies the overlap between predicted and ground truth hand regions. This metric provides a comprehensive measure of segmentation accuracy by comparing the predicted segmentation mask to the ground truth.

To quantify the uncertainty of predictions, we use the average predictive entropy, a standard metric that evaluates the uncertainty associated with each pixel’s prediction across all classes. For a given pixel \(p\), the predictive entropy \(E(p)\) is defined as:

\begin{equation}
E(p) = -\sum_{c=1}^{C} P(c|p) \log P(c|p)
\end{equation}

where \(P(c|p)\) is the predicted probability of class \(c\) for pixel \(p\). Since this is a binary segmentation task (hand and background), the number of classes is \(C=2\). 

For deep ensemble models, the predicted probability \(P(c|p)\) is computed as the average of the output scores from \(K\) base learners:

\begin{equation}
P(c|p) = \frac{1}{K} \sum_{k=1}^{K} P_k(c|p)
\end{equation}

where \(P_k(c|p)\) is the predicted probability from the \(k\)-th base learner. The predictive entropy provides a pixel-level measure of uncertainty, with higher values indicating greater uncertainty in the model's prediction for that pixel.
The average predictive entropy for the entire image, denoted as $\bar{E}$, is calculated as:

\begin{equation}
\bar{E} = \frac{1}{N} \sum_{p=1}^{N} E(p)
\end{equation}

where $N$ is the total number of pixels in the image.

To further evaluate uncertainty specific to the hand regions, we calculate the average predictive entropy for ground truth hand pixels, denoted as $\bar{E}_h$, which focuses only on pixels that belong to the ground truth hand regions. This measure is computed as:

\begin{equation}
\bar{E}_h = \frac{1}{N_h} \sum_{p \in H} E(p)
\end{equation}

where $N_h$ is the total number of ground truth hand pixels, and $H$ represents the set of pixels corresponding to the hand regions in the ground truth. This metric helps evaluate whether the model exhibited uncertainty specifically in predicting the hand regions.

Both metrics are used in this study to evaluate the performance of ID and OOD data by comparing the mIoU scores for accuracy and analyzing the uncertainty of predictions by comparing average predictive entropy to assess the degree of uncertainty exhibited by the models when predicting ID versus OOD data.

\section{Results and discussions}
\label{sec:results}

In this section, we evaluate the performance of the proposed deep ensemble models on human hand segmentation using metrics such as mIoU for segmentation accuracy and average predictive entropy for uncertainty quantification. The evaluations were performed under both ID and OOD conditions. We also present qualitative examples of some ID and OOD examples to discuss the segmentation performances in more detail.

\subsection{Training and validation}

For training, we selected annotated images from each dataset represented in Table \ref{tab:1}. The data was split in a 9:1 ratio for training and validation respectively. It is important to note that the images in all four training datasets are in RGB format, except for HADR, which includes Depth information. For consistency, we utilized only the RGB format of the HADR dataset.
All images were resized and reshaped to ensure compatibility with the generated test dataset.

The deep ensemble architecture (DE-Mix) was trained using the training data. Validation data was used to monitor the mIoU during training. We evaluated the performance of ensemble models with $K = 2, 4, 6, 8, 10$, and observed that model achieved an mIoU exceeding 0.80 on the validation set for $K \geq 4$ across all different training datasets. Based on these results, we selected $K = 4$ for all subsequent evaluations to balance performance and computational efficiency.

\subsection{Quantitative results}

We evaluated the trained models based on our test data, on egocentric images for models trained on EgoHands and Ego2Hands and on side-view images for models trained on HAGS and HADR-trained models. 

\subsubsection{Background effect}
To consider the effect of background in images we compare the segmentation results of pre-trained models on our test images when one operator (O1) is interacting with robot in a simple background (without any other objects) and when the background is cluttered with a lot of industrial objects. The results regarding segmentation accuracy (mIoU) and entropy of predictions of entire image ($\bar{E}$) and hands specifically ($\bar{E}_h$) are listed in Table \ref{tab:3}.

\begin{table*}[]
\caption{Prediction accuracy and predictive uncertainty of the DE-Mix model trained on each dataset using our test images with simple and cluttered backgrounds.}
\centering
\label{tab:3}

\begin{tabular}{lllll}
\toprule
Training dataset           & Test image condition                      & \multicolumn{1}{c}{\begin{tabular}[c]{@{}c@{}}Average\\ mIoU\end{tabular}} & \multicolumn{1}{c}{\begin{tabular}[c]{@{}c@{}}Average\\ $\bar{E}$\end{tabular}} & \multicolumn{1}{c}{\begin{tabular}[c]{@{}c@{}}Average\\ $\bar{E}_h$\end{tabular}} \\ \midrule
\multirow{2}{*}{EgoHands}  & One operator (O1) in simple background    & 0.382                                                                      & 0.114                                                                           & 0.195                                                                             \\
                           & One operator (O1) in cluttered background & 0.361                                                                      & 0.163                                                                           & 0.223                                                                             \\ \midrule
\multirow{2}{*}{Ego2Hands} & One operator (O1) in simple background    & 0.377                                                                      & 0.082                                                                           & 0.212                                                                             \\
                           & One operator (O1) in cluttered background & 0.318                                                                      & 0.140                                                                           & 0.275                                                                             \\ \midrule
\multirow{2}{*}{HADR}      & One operator (O1) in simple background    & 0.419                                                                      & 0.095                                                                           & 0.229                                                                             \\
                           & One operator (O1) in cluttered background & 0.401                                                                      & 0.148                                                                           & 0.261                                                                             \\ \midrule
\multirow{2}{*}{HAGS}      & One operator (O1) in simple background    & 0.496                                                                      & 0.158                                                                           & 0.186                                                                             \\
                           & One operator (O1) in cluttered background & 0.479                                                                      & 0.195                                                                           & 0.283                                                                             \\ 
                           \bottomrule
\end{tabular}
\end{table*}

As shown in Table \ref{tab:3}, the segmentation accuracy (mIoU) for models trained on EgoHands and Ego2Hands is significantly lower compared to models trained on HADR and HAGS, regardless of background conditions. This discrepancy can be attributed to the non-industrial nature of the EgoHands and Ego2Hands datasets. The presence of industrial robots and workplace elements in the test images introduces challenges for these models, as their training data lacks such scenarios, reducing their generalizability to segment hands in industrial settings.
In contrast, the model trained on HAGS outperforms the model trained on HADR, which may be due to the synthetic nature of the HADR dataset. Models trained on synthetic data often struggle with domain adaptation when applied to real-world scenarios. Additionally, the HADR dataset primarily contains images captured from a top-down view, whereas HAGS includes both top-down and side-view images, making it more similar to our test dataset and therefore better suited for segmentation in these conditions.
Overall, segmentation accuracy decreased for all models when tested in cluttered and messy backgrounds compared to simple and minimalist backgrounds. This drop is more pronounced for models trained on egocentric datasets (EgoHands and Ego2Hands), likely due to the absence of industrial tools and environments in the backgrounds of their training data.

Examining the average predictive entropy of the entire image ($\bar{E}$) reveals that simpler backgrounds consistently result in lower entropy compared to cluttered backgrounds across all pre-trained models. This indicates that models are more confident in their predictions when the background is less complex while when the background is messy it increases the uncertainty of the model so that it struggles to classify each pixel to hand or background.
Focusing on the pixels corresponding to hand regions, the average predictive entropy for hands ($\bar{E}_h$) is substantially higher than the overall entropy ($\bar{E}$), suggesting that models are inherently less confident when predicting hand pixels. This uncertainty becomes even more pronounced in cluttered backgrounds, likely because hands are often in close proximity to other objects, making it more challenging for models to confidently classify those pixels as hands. 

\subsubsection{ID and OOD data}

The segmentation performance (mIoU) and predictive entropy for the deep ensemble model (DE-Mix) trained on four training datasets are listed in Table \ref{tab:4} for both in-distribution (ID) and out-of-distribution (OOD) data. Notably, all test conditions involve messy and cluttered backgrounds with industrial tools, ensuring realistic and challenging evaluation scenarios. To provide a structured analysis, the discussion is divided into two parts: segmentation accuracy (mIoU) and predictive entropy.

\begin{table*}[]
\caption{Segementation performance and predictive uncertainty of models trained on each dataset using our test data in ID and OOD conditions}
\resizebox{\textwidth}{!}{
\centering
\label{tab:4}

\begin{tabular}{llllll}
\toprule
Training dataset           & ID / OOD             & Image condition in test phase                                 & \multicolumn{1}{c}{\begin{tabular}[c]{@{}c@{}}Average\\ mIoU\end{tabular}} & \multicolumn{1}{c}{\begin{tabular}[c]{@{}c@{}}Average\\ $\bar{E}$\end{tabular}} & \multicolumn{1}{c}{\begin{tabular}[c]{@{}c@{}}Average\\ $\bar{E}_h$\end{tabular}} \\ \midrule
\multirow{7}{*}{EgoHands}  & \multirow{2}{*}{ID}  & One operator (O1)                                             & 0.361                                                                      & 0.163                                                                           & 0.223                                                                             \\
                           &                      & Two operators (O2)                                            & 0.306                                                                      & 0.172                                                                           & 0.278                                                                             \\ \cline{2-6} 
                           & \multirow{5}{*}{OOD} & One operator (O1) with gloves (GH)                            & 0.184                                                                      & 0.157                                                                           & 0.211                                                                             \\
                           &                      & Two operators (O2) with gloves (GH)                           & 0.115                                                                      & 0.174                                                                           & 0.215                                                                             \\
                           &                      & Two operators (O2) having rare gestures (RG)                  & 0.052                                                                      & 0.168                                                                           & 0.231                                                                             \\
                           &                      & Two operators (O2) with gloves (GH) having rare gestures (RG) & 0.058                                                                      & 0.170                                                                           & 0.164                                                                             \\
                           &                      & Images with motion blurred noise (MBN)                        & 0.103                                                                      & 0.154                                                                           & 0.126                                                                             \\ \midrule
\multirow{7}{*}{Ego2Hands} & ID                   & One operator (O1)                                             & 0.308                                                                      & 0.140                                                                           & 0.275                                                                             \\ \cline{2-6} 
                           & \multirow{6}{*}{OOD} & Two operators (O2)                                            & 0.202                                                                      & 0.153                                                                           & 0.247                                                                             \\
                           &                      & One operator (O1) with gloves (GH)                            & 0.171                                                                      & 0.134                                                                           & 0.198                                                                             \\
                           &                      & Two operators (O2) with gloves (GH)                           & 0.105                                                                      & 0.165                                                                           & 0.250                                                                             \\
                           &                      & Two operators (O2) having rare gestures (RG)                  & 0.069                                                                      & 0.146                                                                           & 0.212                                                                             \\
                           &                      & Two operators (O2) with gloves (GH) having rare gestures (RG) & 0.033                                                                      & 0.162                                                                           & 0.298                                                                             \\
                           &                      & Images with motion blurred noise (MBN)                        & 0.115                                                                      & 0.149                                                                           & 0.165                                                                             \\ \midrule
\multirow{7}{*}{HADR}      & ID                   & One operator (O1)                                             & 0.419                                                                      & 0.148                                                                           & 0.261                                                                             \\ \cline{2-6} 
                           & \multirow{6}{*}{OOD} & Two operators (O2)                                            & 0.349                                                                      & 0.155                                                                           & 0.312                                                                             \\
                           &                      & One operator (O1) with gloves (GH)                            & 0.279                                                                      & 0.131                                                                           & 0.329                                                                             \\
                           &                      & Two operators (O2) with gloves (GH)                           & 0.201                                                                      & 0.160                                                                           & 0.412                                                                             \\
                           &                      & Two operators (O2) having rare gestures (RG)                  & 0.148                                                                      & 0.152                                                                           & 0.403                                                                             \\
                           &                      & Two operators (O2) with gloves (GH) having rare gestures (RG) & 0.123                                                                      & 0.150                                                                           & 0.365                                                                             \\
                           &                      & Images with motion blurred noise (MBN)                        & 0.239                                                                      & 0.135                                                                           & 0.287                                                                             \\ \midrule
\multirow{7}{*}{HAGS}      & \multirow{2}{*}{ID}  & One operator (O1)                                             & 0.479                                                                      & 0.195                                                                           & 0.283                                                                             \\
                           &                      & One operator (O1) with gloves (GH)                            & 0.465                                                                      & 0.182                                                                           & 0.272                                                                             \\ \cline{2-6} 
                           & \multirow{5}{*}{OOD} & Two operators (O2)                                            & 0.362                                                                      & 0.204                                                                           & 0.348                                                                             \\
                           &                      & Two operators (O2) with gloves (GH)                           & 0.346                                                                      & 0.199                                                                           & 0.302                                                                             \\
                           &                      & Two operators (O2) having rare gestures (RG)                  & 0.238                                                                      & 0.196                                                                           & 0.417                                                                             \\
                           &                      & Two operators (O2) with gloves (GH) having rare gestures (RG) & 0.229                                                                      & 0.201                                                                           & 0.404                                                                             \\
                           &                      & Images with motion blurred noise (MBN)                        & 0.278                                                                      & 0.193                                                                           & 0.292                                                                             \\ \bottomrule
\end{tabular}
}
\end{table*}

eaning when human hands are in the movement models struggle segmenting the hands pixels.

\textbf{Accuracy of segmentation (mIoU):} As can be seen from Table \ref{tab:4}, the trained model has better segmentation accuracy (mIoU) for the conditions it encountered during training (ID data) compared to the scenarios it faced without prior exposure (OOD data) across all four training datasets. 
The model trained on EgoHands when segment hands of one or two human operators in the scene (O1 and O2), it maintains segmentation accuracy for two operators (ID data) since this condition was present during training. In contrast, the model trained on Ego2Hands experiences a significant accuracy drop in the same scenario since it has not faced two humans in its training (OOD data). 
In other OOD scenarios (e.g., two operators with and without gloves), models trained on these egocentric datasets perform poorly, with mIoU values below 0.2. Particularly for rare gestures (RG), these models fail to segment even a small portion of the operators' hands (mIoU less than 0.1). Similarly, for motion-blurred images (MBN), representing aleatoric uncertainty, these models achieve an mIoU around 0.1, effectively failing to segment hands.
On the other hand, models trained on HADR and HAGS datasets show significantly better segmentation accuracy in both ID and OOD data. This performance can be attributed to their industrial context, which closely matches the testing conditions in this study. The model trained on HAGS performs better than the one trained on HADR, likely because HAGS includes gloved hands (GH) in its training data and contains both top-down and side-view perspectives, which are similar to our test dataset.
In O2 scenarios (two operators), both HAGS- and HADR-trained models experience an accuracy drop, as these conditions are OOD for their training sets. However, the HAGS-trained model achieves better mIoU (~0.35) compared to the HADR-trained model. In rare gesture (RG) and motion-blurred (MBN) scenarios, both models struggle, but the HAGS-trained model performs slightly better (mIoU ~0.23 for RG, ~0.28 for MBN) compared to HADR (mIoU~0.12 for RG, ~0.24 for MBN).

\textbf{Predictive Entropy (\(\bar{E}\), \(\bar{E}_h\)):} From Table \ref{tab:4}, it can be observed that the average predictive entropy for the entire image (\(\bar{E}\)) remains relatively stable between ID and OOD data. This stability contrasts with the significant change observed in Table \ref{tab:3} when comparing simple and cluttered backgrounds. This is likely because hand regions constitute a small proportion of the entire image, so changes in hand conditions do not significantly affect the overall uncertainty of the image.
However, when examining the predictive entropy for hand pixels (\(\bar{E}_h\)), a good model is expected to exhibit higher uncertainty for OOD data compared to ID data. Models trained on EgoHands and Ego2Hands fail to follow this pattern. Their low mIoU values suggest that these models incorrectly classify hand pixels as background with high confidence, making them unreliable for segmenting hands in OOD conditions.
In contrast, models trained on HAGS and HADR report higher predictive entropy (\(\bar{E}_h\)) for hand regions in OOD conditions, indicating appropriate recognition of unfamiliar scenarios as uncertain. This behavior aligns with the expected characteristics of robust models.

The superiority of models trained on HAGS and HADR is clearly evident compared to those trained on EgoHands and Ego2Hands in terms of both segmentation accuracy and predictive entropy. This can be attributed to the fact that the images in HAGS and HADR datasets are more aligned with our testing conditions, as both datasets are specifically designed for industrial human-robot interaction scenarios. In contrast, the images in EgoHands and Ego2Hands primarily depict hands in general, non-industrial activities, making them less relevant for the industrial context of our evaluation. 

It is important to note that the results reported above are based on models evaluated directly without fine-tuning on the test dataset. This deliberate choice ensures that the evaluation highlights the models' generalization capabilities in novel, realistic industrial conditions. However, this approach inherently results in lower segmentation accuracy (mIoU) compared to models that undergo fine-tuning on similar test data.

All experiments were conducted on a workstation equipped with an NVIDIA GeForce RTX 3090 GPU, an AMD Ryzen 9 processor, and 64 GB of memory. This setup provided the computational power required for training and evaluating the deep ensemble model (DE-Mix). With this configuration, the DE-Mix model (with $K = 4$ base learners) achieved a segmentation speed of 38 frames per second, demonstrating its capability for real-time performance in practical applications.

\subsection{Qualitative results}

The segmentation results for both ID and OOD images captured from egocentric and side cameras are presented in this section. 

\begin{figure*}[ht]
 \centering
 \includegraphics[width=14cm]{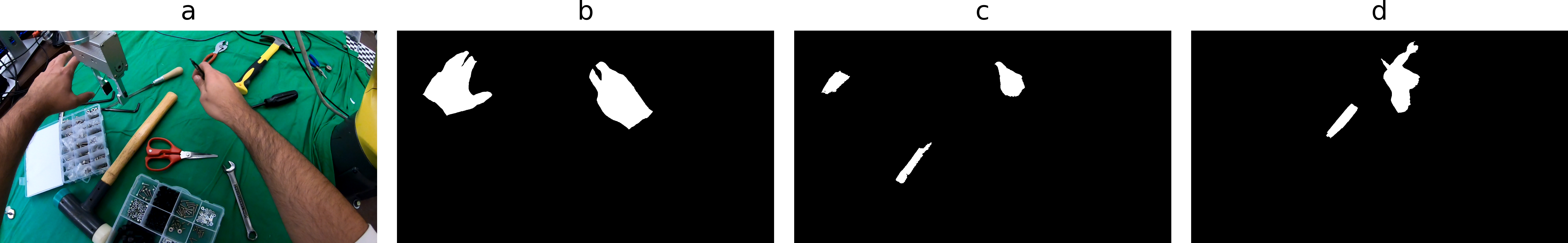}
 \caption{a) an ID data in the egocentric view, b) the segmentation ground truth mask, c) segmented by DE-Mix trained on EgoHands, d) segmented by DE-Mix trained on Ego2Hands.}
 \label{fig:5}
\end{figure*}

Figure \ref{fig:5} shows an egocentric image of ID data, depicting one operator (O1) interacting with the environment. The ground truth and predicted segmentation masks from models trained on EgoHands and Ego2Hands are displayed. The model trained on EgoHands is able to segment some pixels of both hands of the operator but mistakenly identifies part of the wooden handle of the hammer as hand pixels. In contrast, the model trained on Ego2Hands segments only a small portion of one hand and mistakenly assigns hand labels to the wooden handle of the hammer and another nearby tool. These results indicate the limited accuracy of models trained on these datasets even dealing with ID data.

\begin{figure*}[ht]
 \centering
 \includegraphics[width=14cm]{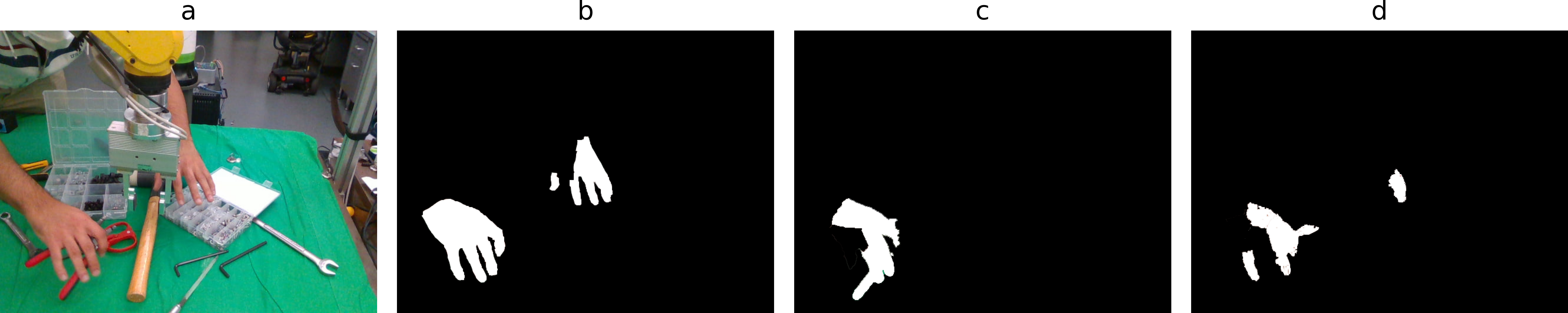}
 \caption{a) an ID data in the side view, b) the segmentation ground truth mask, c) segmented by DE-Mix trained on HADR, d) segmented by DE-Mix trained on HAGS.}
 \label{fig:6}
\end{figure*}

Figure \ref{fig:6} depicts a side-view image of the same condition shown in Figure \ref{fig:5}, with segmentation results from models trained on HADR and HAGS. The model trained on HADR segments some pixels of one hand while misclassifying parts of the tool carried by the operator as hand pixels. Conversely, the model trained on HAGS demonstrates better performance by detecting portions of both hands, even though the operator's left hand is occluded by the industrial robot. This highlights the higher accuracy of the HAGS-trained model compared to the HADR-trained model in ID scenarios.

\begin{figure*}[ht]
 \centering
 \includegraphics[width=14cm]{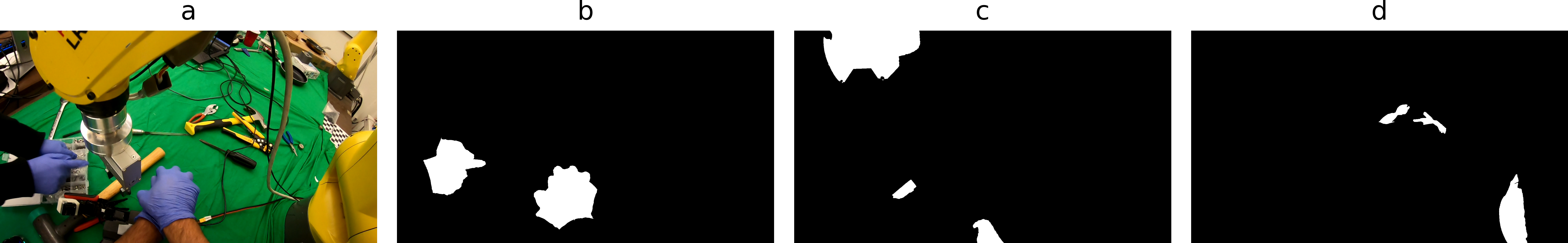}
 \caption{a) an OOD data of two human operators O2 with gloves (GH) having rare gestures(RG) in the he egocentric view, b) the segmentation ground truth mask, c) segmented by DE-Mix trained on EgoHands, d) segmented by DE-Mix trained on Ego2Hands.}
 \label{fig:7}
\end{figure*}

Figure \ref{fig:7} illustrates an OOD image featuring two operators (O2) wearing gloves (GH) and performing rare gestures (RG). The segmentation masks from the model trained on EgoHands reveal segmenting only a few pixels of the operators’ forearms, which are not considered part of the hand since training and testing data define hands as the region from the fingertips to the wrist. Additionally, the model incorrectly identifies parts of the robotic arm and the wooden hammer handle as hand pixels. The model trained on Ego2Hands, however, fails to segment any human hand pixels while erroneously assigning hand labels to parts of the robotic arm and tools. These observations explain the near-zero mIoU values reported for these models when dealing with these kinds of OOD data, as shown in Table \ref{tab:4}.

\begin{figure*}[ht]
 \centering
 \includegraphics[width=14cm]{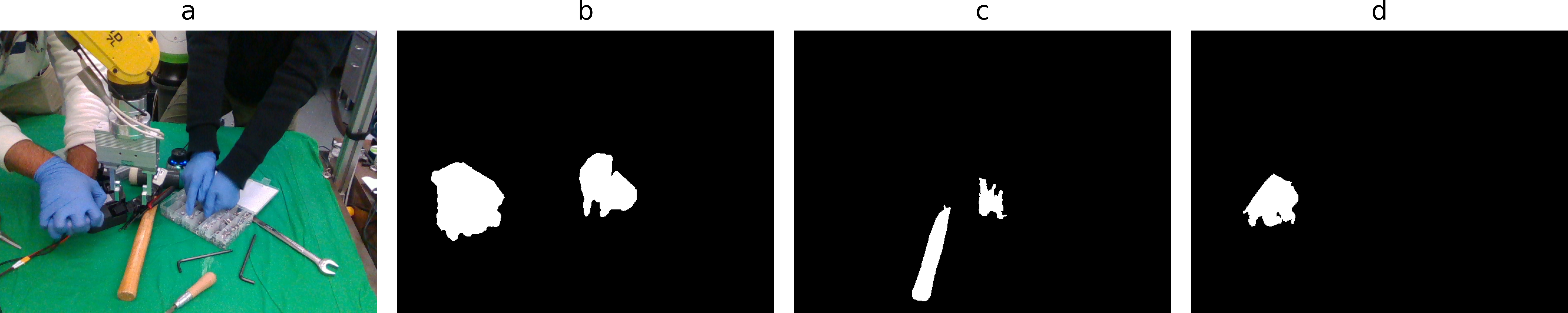}
 \caption{a) an OOD data of two human operators O2 with gloves (GH) having rare gestures(RG) in the side view, b) the segmentation ground truth mask, c) segmented by DE-Mix trained on HADR, d) segmented by DE-Mix trained on HAGS.}
 \label{fig:8}
\end{figure*}

Figure \ref{fig:8} displays the side-view perspective of the same OOD scenario shown in Figure \ref{fig:7}. The model trained on HADR correctly segments a small portion of one operator’s hand but misclassifies the wooden hammer handle as hand pixels. The model trained on HAGS, however, accurately segments portions of one operator’s hand without incorrectly assigning other pixels as hands, although it fails to detect the other operator's hands. These results align with the lower mIoU values observed for OOD data compared to ID data, as reported in Table \ref{tab:4}.

\section{Conclusions and future works}
\label{sec:conclusions}

This study investigated the performance of a deep ensemble model, DE-Mix, composed of UNet and RefineNet, for human hand segmentation in industrial human-robot interaction scenarios. Using four datasets—EgoHands, Ego2Hands, HADR, and HAGS—we evaluated the generalization capabilities of models trained on these datasets under diverse and challenging conditions, including both in-distribution (ID) and out-of-distribution (OOD) scenarios. The results underscored the importance of domain-specific datasets, with models trained on HAGS and HADR consistently outperforming those trained on EgoHands and Ego2Hands in segmentation accuracy and predictive uncertainty metrics.

Although OOD conditions posed significant challenges for all models, those trained on industrial datasets demonstrated greater robustness in handling scenarios such as gloved hands, rare gestures, and motion blur. In contrast, models trained on non-industrial datasets struggled to generalize, often failing to accurately segment hands or misclassifying non-hand regions. The complex and realistic nature of the test dataset characterized by cluttered backgrounds, industrial tools, and multiple human hands highlighted the inherent challenges of deploying segmentation models in dynamic, real-world environments. While segmentation accuracy was relatively low due to the absence of fine-tuning on the test data, this limitation emphasized the challenges of generalizing to novel industrial conditions, reinforcing the need for more diverse datasets and adaptive techniques.

Future research could focus on fine-tuning pre-trained models with industrial datasets to enhance their applicability to specific scenarios. Expanding the diversity of training datasets by incorporating rare gestures, motion blur, and occluded hands could be another option although constructing such comprehensive datasets is a resource-intensive and time-consuming endeavor. We also restrict our experience to two human operators and also two cameras, i.e. egocentric and side camera, extending future work to include more complex scenarios, such as multiple operators and varied camera perspectives such as top-down camera, would provide further insights. Additionally, integrating multimodal data, such as depth, may further enhance segmentation accuracy under challenging conditions.

Exploration of advanced model architectures, such as Transformers, offers promising potential to improve spatial understanding and generalization to OOD scenarios. Furthermore, optimizing lightweight models for real-time inference would facilitate deployment in industrial environments, where speed and efficiency are critical.

\bibliography{biblio}

\end{document}